\documentclass{ewic}
\usepackage{comment}
\usepackage{booktabs} % For formal tables
\usepackage{booktabs} % For formal tables
\usepackage{acronym}
\usepackage{booktabs} % For formal tables
\usepackage{amsfonts}
\usepackage{amssymb}
\usepackage{physics}
\usepackage{braket}
\usepackage{subfigure}
\usepackage{acronym}
\usepackage{amsmath}
\usepackage{url}

\acrodef{ML}[ML]{Machine Learning}
\acrodef{QT}[QT]{Quantum Theory}
\acrodef{PCA}[PCA]{Principal Component Analysis}
\acrodef{IR}[IR]{Information Retrieval}
\acrodef{ANN}[ANN]{Artificial Neural Networks}
\acrodef{QM}[QM]{Quantum Mechanics}
\acrodef{QDT}[QDT]{Quantum Detection Theory}
\acrodef{SDT}[SDT]{Signal Detection Theory}
\acrodef{NB}[NB]{Na\"{\i}ve Bayes}
\acrodef{SVM}[SVM]{Support Vector Machine}
\acrodef{KNN}[k-NN]{$k$ Nearest Neighbours}
\acrodef{DT}[DT]{Decision Tree}
\begin{document}

\runningheads{P.Tiwari, M. Melucci}{Multi-class Classification Model Inspired by Quantum Detection Theory}

\conference{Proceedings of the 8th Symposium on\\
Future Directions in Information Access 2018}

\title{Multi-class Classification Model Inspired by Quantum Detection Theory}

\authorone{Prayag Tiwari, Massimo Melucci\\
Department of Information Engineering\\
University of Padova\\
Padova, Italy\\
\email{tiwari@dei.unipd.it, melo@dei.unipd.it}}

\begin{abstract}
Machine Learning has become very famous in today's world  which assist to identifying the patterns from the raw data. Technological advancement has led to substantial improvement in Machine Learning which, thus helping to improve prediction. Current Machine Learning models are based on Classical Theory, which can be replaced by Quantum Theory to improve the effectiveness of the model.

In the previous work, we developed binary classifier inspired by Quantum Detection Theory. In this extended abstract, our main goal is to develop multi class classifier. We generally use the terminology multinomial classification or multi-class classification when we have a classification problem for classifying observations or instances into one of three or more classes. 
%This research aims to address a newly important issues of information behaviour research by reviewing the approaches to information seeking and retrieval in the context of Information Foraging Theory.
\end{abstract}


\maketitle

\section{Introduction}
Quantum Mechanics has already shown its effectiveness in many fields so there is a good possibility that it will prove to be useful in \ac{ML} as well. Quantum theory can open a new way towards quantum inspired \ac{ML} which might outperform traditional machine learning if used properly. The theory of \ac{QM} has been implemented in several domain of IR recently  by~\cite{li2018quantum,zhang2018quantum,zhang2016quantum,zhang2018end,wang2016exploration,zhang2018quantum1,li2018quantum,li2015modeling} . Quantum Probability theory is the quantum generalization of classical probability theory, which was developed by \cite{vonNeumann55}. Classical probability theory provides that a system can have either state 0 or 1 and quantum probability comes into existence to go beyond classical theory and describe states in between 0 and 1 with classical states.

The effectiveness of the state-of-the-art classification algorithms
relies on logical theory of sets, theory of probability and the
algebra of vector spaces.  For example, the most straightforward
technique is called Naive Bayes, which considers objects (e.g. documents)
as elements of sets and applies basic probability measures to these
sets for selecting classes.  Another effective classification
technique called Support Vector Machines considers objects as points
of a multi-dimensional space and aims to select subspaces as classes.
However, an effective combination of techniques stemming from
different theories is still missing, although it has been investigated in IR
since the book on the Geometry of IR by \cite{vanRijsbergen79a}.

Despite its effectiveness in some domains, classification
effectiveness is still unsatisfactory in a number of application
domains due to a variety of reasons, such as
the number of categories and the nature of data.  The
number of categories may be so large that a classification
technique that is effective for a few categories may be ineffective when
thousands of categories are required; moreover, the nature of data may be
so complex that the techniques that are effective for simple objects
may prove to be ineffective for complex objects.  A sensible approach to addressing the problems
caused by unconventional categorical systems or complex data is to
adapt well-known and effective techniques to these new contexts.
Another approach, which is indeed the focus of this paper, is to
radically change paradigm and to investigate whether a new theoretical
framework may be beneficial and be a new research direction.

Quantum Theory may provide a theoretical model for classification. To our knowledge no work has been done on quantum based classification so this paper is a first step
 to enter into quantum inspired machine learning and prove the effectiveness of quantum theory in  
classification.

\section{Proposed Methodology and Discussion}
In the previous work, we developed a binary classifier inspired by Quantum Detection Theory by \cite{di2018binary}. The main task was to identify whether a document belongs to a given topic or not. We used Reuters21578\footnote{\url{http://www.daviddlewis.com/resources/testcollections/reuters21578/}} in order to check the performance of our model. Our proposed binary classifier model inspired from quantum detection theory performed very well in terms of recall and  F-measures for most of the topics. An experiment was done on the small dataset so  work is still in progress. Our algorithm works as follows: it starts by computing the density operators $\rho_1$ and $\rho_0$ from
positive  and negative samples, respectively. In order to achieve this, for a particular feature, we
first compute the number of documents with non-zero values in the
feature. In this way, one vector is
generated from each class, thus obtaining two vectors which are
respectively denoted as $\vert v_1 \rangle$ and $\vert v_0
\rangle$. These vectors can be regarded as a representation of the
feature statistics among a class. We normalize the vectors and compute the outer spaces in order to obtain the density operators $\rho_1$ and $\rho_0$:

\begin{equation}
\label{eq:rhos}
\rho_1  = \frac{\vert v_1 \rangle \langle v_1 \vert}{tr(\vert v_1 \rangle \langle v_1 \vert)} \qquad
\rho_0 = \frac{\vert v_0 \rangle \langle v_0 \vert}{tr(\vert v_0 \rangle \langle v_0 \vert)}
\end{equation}
We computed the projection operator $P$ according to the eigen        decomposition described in
\citep{Melucci15b}, that is,
\begin{equation}
  \label{eq:decomposition}
  \rho_1 - \lambda \rho_0 = \eta\, P + \beta\,P^\perp \qquad \eta > 0
  \qquad \beta < 0 \qquad P\,P^\perp = 0
\end{equation}
where $\xi$ is the prior probability of the negative class and $
\lambda = {\xi}\,/\,(1-\xi) $. Moreover, $\eta$ is the positive
eigenvalue corresponding to $P$ which represents the subspaces of the
vectors representing the documents to be accepted in the target class.

In this extended abstract, our main goal is to develop multi class classifier. We generally use the terminology "multinomial classification or multi-class classification" when we have a classification problem for classifying observations or instances into one of three or more classes. 

The main theory behind quantum inspired multi-class classification is as follows: The choice among the $N$ hypotheses, which the $k^{th}$ asserts "The system has the density operator $\rho_k$," in which $k =1, 2, 3, ....., N$ can be based on the result of the measurement of $N$ commuting operators $P_1, P_2, ...., P_N$, making a resolution of identity operator $1$:

\begin{equation}
P_1+P_2+ ....+P_N = 1
\end{equation}

Our problem is getting the set of projectors so that the choice among the $N$ hypothesis can be made with the minimum average cost. It will assist in the event of Quantum Detection Theory, in constructing and estimating the best receiver for the communications system. In this, messages are coded into an alphabets of 3 or more symbols, and a distinct signal is transmitted for each.

Assume $\xi_k$ is the prior probability of the hypothesis $H_k$, and $K_{ij}$ is the cost for choosing $H_i$ when $H_j$ is correct. So the average cost per decision can be described as,
\begin{equation}
\bar{K} = \sum^N _{i=1} \sum^N _{j=1} \xi_j K_{ij} \mbox Tr(\rho_j P_i) ,
\end{equation}
 which has to be minimized by the set of given commuting projection operators $P_k$. In particular, $K_{ii}=0$, $K_{ij}=1$, $i\neq j$, $\bar{k}$ can be approximate to the average probability of error.
 
In each hypotheses, the state of the system is in pure state $\rho_k = \vert \psi_k \rangle \langle \psi_k \vert$, so the projection operator will have such form, $P_j =\vert \eta_j \rangle \langle \eta_j \vert$. Here $\vert \eta_j \rangle$ is the linear combination of the the given state $\vert \psi_k \rangle$. The main problem is to find the  set of projectors minimizing the average cost when more than two  
categories or hypotheses are available; the solution can be a  
generalization of the solution of the problem of finding the set of  
projectors in the event of two categories.

\section{Conclusion and Future Works}
Although research work is still in progress  
we are testing a multi class classifier based on Quantum Detection  
Theory and we expect that it is possible to develop such a model. In order to learn about \ac{QDT} and classification tasks in more detail, these works may be beneficial and a basis for developing this model. \citep{helstrom1971quantum,yuen1975optimum,helstrom1969quantum,helstrom1972vii,eldar2001quantum,helstrom1974noncommuting,helstrom1968detection,holevo1998capacity,vilnrotter2001quantum,di2016evaluation,melucci2015introduction,melucci2012contextual,melucci2011quantum,melucci2011advanced,melucci2018efficient,melucci2017algorithm,nanni2016combination} 

\section*{Acknowledgement}
This work is supported by the Quantum Access and Retrieval Theory (QUARTZ) project, which has received funding from the European Union’s Horizon 2020 research and innovation programme under the Marie Sklodowska-Curie grant agreement No 721321.

\bibliography{refs}{}
\bibliographystyle{agsm}
\begin{comment}

\end{comment}
\end{document}